\documentclass[onecolumn,preprint]{article}

\usepackage{graphicx}          
\usepackage{parskip}
\usepackage[hidelinks]{hyperref}
\usepackage{amsmath,amssymb,multirow,amsfonts}
\usepackage{siunitx}
\usepackage{booktabs}
\usepackage{algorithm,algpseudocode}
\newcommand{\myd}{\textrm{d}}      
\newcommand{\Transp}{\mathsf{T}}
\newcommand{\norm}[1]{\left\lVert#1\right\rVert}
\newcommand{\erf}{\text{erf}}
\newcommand{\Npdf}[3]{\mathcal{N}\left(#1 | #2,#3\right)}
\newcommand{\pmat}[1]{\begin{bmatrix}#1\end{bmatrix}}
\usepackage[affil-it]{authblk}
\usepackage[margin=2.5cm]{geometry}

\date{}

\title{Evaluating the squared-exponential covariance function in Gaussian processes with integral observations} 

\author[1]{J.N. Hendriks}
\author[2]{C. Jidling}               
\author[1]{A. Wills}  
\author[2]{T. B. Sch\"{o}n}

\affil[1]{School of Engineering, University of Newcastle, Australia.}  
\affil[2]{Department of Information Technology, Uppsala University, Sweden.}

\begin{document}
  \maketitle
\begin{abstract}                          
This paper deals with the evaluation of double line integrals of the squared exponential covariance function. We propose a new approach in which the double integral is reduced to a single integral using the error function. This single integral is then computed with efficiently implemented numerical techniques. The performance is compared against existing state of the art methods and the results show superior properties in numerical robustness and accuracy per computation time.
\end{abstract}


\section{Introduction} 
\label{sec:introduction}
The Gaussian process (GP) \cite{rasmussen2006gaussian} has over the past decade become an important and fairly standard tool for system identification. 
The flexibility of the GP and more specifically its inherent data-driven bias-variance trade off makes it a very successful model for linear systems \cite{PillonettoDN:2010,ChenOL:2012,PillonettoDCNL:2014}, where it in fact improves performance compared to more classical models. 
When it comes to nonlinear system the GP has also shown great promise, for example when combined with the state space model \cite{FrigolaLSR:2013,FrigolaCR:2014,SvenssonS:2017} and when used in a classic auto-regressive setting \cite{KocijanGBMS:2005,BijlSWV2017}. 

A useful property of GP is that it is closed under linear operators, such as integrals  \cite{rasmussen2006gaussian,papoulis1965probability,graepel2003solving}.
This enables GP regression to be applied to data that is related to the underlying object of interest via a \textit{line integral}. 
In this case, evaluating the resulting measurement covariance involves double line integrals of the covariance function. There are no analytical solutions available and we are forced to numerical approximations. 
To give a few examples of where this problem arises, we mention; optimization algorithms \cite{hennig2012quasi,wills2017construction}, continuous-time system identification, quadrature rules \cite{minka2000deriving}, and  tomographic reconstruction \cite{jidling2018probabilistic}. 

The \textit{contribution} of this paper is a numerically efficient and robust method for solving these integrals when the squared-exponential covariance function is used. We also provide a thorough comparison of the accuracy and computational time to existing state of the art methods for this rather specific problem. 

\section{Problem Statement} 
\label{sec:problem_statement}
Consider a set of $n$ line integral observations of the form 
\begin{equation}
y_i = \lvert\lvert\mathbf{w}_i\rvert\rvert\int_0^1 f(\mathbf{w}_i s+\mathbf{p}_i)\,\mathrm{d}s + e_i,
\end{equation}
that provide measurements of a scalar function $f$ over the input space $\mathbf{z}\in \mathbb{R}^m$, where $e_i\sim\mathcal{N}(0,\sigma_n^2)$\footnote{The notation denotes that $e_i$ is normally distributed with mean $0$ and variance $\sigma_n^2$.}. 
By $\mathbf{p}_i$ we denote the start point of the line while the vector $\mathbf{w}_i$ specifies its direction and length; hence the end point is given by $\mathbf{w}_i+\mathbf{p}_i$, and any point on the line can be expressed as $\mathbf{w}_i s+\mathbf{p}_i$ with $s\in[0,1]$.
Given this set we wish to make predictions $f_*$ of the function for a new input $\mathbf{z}_*$.


In GP regression this is accomplished through the specification of a prior covariance function $k(\mathbf{z},\mathbf{z}')$ which defines the correlation between any two function values and loosely speaking controls how smooth we believe the underlying function to be.
Although there exists many possible covariance functions \cite{rasmussen2006gaussian}, here we will consider only the squared-exponential covariance function;
\begin{equation}
    k(\mathbf{z},\mathbf{z}') = \exp\left(-0.5(\mathbf{z}-\mathbf{z}')^T\mathbf{V}(\mathbf{z}-\mathbf{z}')\right),
\end{equation}
where $\mathbf{V}$ is a positive definite and symmetric scaling matrix.

The covariance between a measurement $y_i$ and the function at a test input $\mathbf{z}_*$ is given by the single integral
\begin{equation}\label{eq:singleInt}
    \mathbf{K}_{i*} = \norm{\mathbf{w}_i}\int\limits_0^1 \exp\left(-0.5(\mathbf{w}_i t + \mathbf{v}_{i})^T\mathbf{V}(\mathbf{w}_i t + \mathbf{v}_i)\right)\,\mathrm{d}t,
\end{equation}
where $\mathbf{v}_i = \mathbf{p}_i - \mathbf{z}_*$. Two cases need considering.

\textbf{Case 1:} $\norm{\mathbf{w}_i} = 0$. Then the integral \eqref{eq:singleInt} reduces to 
\begin{equation} \label{eq:solsingle1}
  \mathbf{K}_{i*} = \norm{\mathbf{w}_i}\exp\left(-\frac{1}{2}\mathbf{v}_i^T\mathbf{V}\mathbf{v}_i\right) = 0.
\end{equation}
\textbf{Case 2:} $\norm{\mathbf{w}_i} > 0$. Then the solution is given by
\small
\begin{subequations}\label{eq:solsingle2}
\begin{equation}
\begin{split}
      \mathbf{K}_{i*} &= \norm{\mathbf{w}_i}\sqrt{\frac{\pi}{2c_1}}\exp\left(\frac{c_2^2}{8c_1}-\frac{1}{2}c_3\right)\left(\text{erf}\left(\frac{2c_1+c_2}{2\sqrt{2c_1}}\right)-\text{erf}\left(\frac{c_2}{2\sqrt{2c_1}}\right)\right),
\end{split}
\end{equation}
\normalsize
where $\erf(\cdot)$ is the error function \cite{andrews1992special} and
\begin{equation}
  \begin{split}
    c_1 &= \mathbf{w}_i^T\mathbf{V}\mathbf{w}_i,\quad c_2 = 2\mathbf{w}_i^T\mathbf{V}\mathbf{v}_i,\quad c_3 = \mathbf{v}_i^T\mathbf{V}\mathbf{v}_i.
  \end{split}
\end{equation}
\end{subequations}


Determining the covariance function between a pair of line integral measurements requires evaluating the following double integral
\begin{equation}\label{eq:doubleInt}
\begin{split}
    \mathbf{K}_{i,j} \hspace{-0.75mm}=\hspace{-0.75mm} \norm{\mathbf{w}_i}\norm{\mathbf{w}_j}\hspace{-0.75mm}\int\limits_0^1\hspace{-1.5mm}\int\limits_0^1 \hspace{-0.75mm}\exp\Big(-\frac{1}{2}\left(\mathbf{u}_{ij}-s\mathbf{w}_j+t\mathbf{w}_i\right)^T\mathbf{V}\left(\mathbf{u}_{ij}-s\mathbf{w}_j+t\mathbf{w}_i\right)\Big)\,\mathrm{d}t\mathrm{d}s,
\end{split}
\end{equation}
where $\mathbf{u}_{ij} = \mathbf{p}_i-\mathbf{p}_j$. This is analytically intractable and hence we are forced to numerical approximations.


\section{Related Work}\label{sec:existing_approaches}
Here, we provide a brief overview of existing approaches and their benefits and disadvantages. 

The need to perform a double integral can be avoided entirely by the use of reduced rank approximation schemes that approximate the squared-exponential by a finite set of basis functions. 
One such scheme is the Hilbert space approximation \cite{solin2014hilbert} that was used in \cite{jidling2018probabilistic} to perform inference of strain fields from line integral measurements. 
Although the problem of calculating the double line integral is removed, the number of basis functions required and the subsequent computational cost grows exponentially with the problem dimension; making this method infeasible for higher-dimensional problems.
An additional drawback is that the expressiveness of the model is limited to the basis functions chosen which may make this approach undesirable in many applications.
Although this approach may be appropriate for some applications, the rest of this paper focuses on solutions that do not approximate the covariance function.

A simple approach to evaluate the double line integral is the use of 2D numerical integration methods (e.g. using a 2D Simpson's rule). 
Such an approach has a trade off between computation time and accuracy. 
Although the approach detailed later still requires the numerical evaluation of a single integral, we have found it to be more efficient for the accuracy provided.

Another approach is to transform the problem into the double integral of a bivariate normal distribution. The integral in \eqref{eq:doubleInt} can be rewritten as
\begin{subequations}
 \begin{align}
   I_2 = \int\limits_{0}^{1}\hspace{-1.5mm} \int\limits_{0}^{1} e^{-\frac{1}{2}f(t, s)}
   \,\mathrm{d} t \mathrm{d} s,
 \end{align}
 where the function $f(t, s)$ is given by
 \begin{align}
   \notag
   f(t, s) &= (\mathbf{p}_i - \mathbf{p}_j + \mathbf{w}_{i}t- \mathbf{w}_{j}s)^\Transp \mathbf{V}\\
   &\hspace{20mm}(\mathbf{p}_i - \mathbf{p}_j + \mathbf{w}_{i}t - \mathbf{w}_{j}s)\\
      &= \begin{pmatrix}
        t\\
        s
      \end{pmatrix}^{\Transp}
   \begin{pmatrix}
     a & b\\
     b & c
   \end{pmatrix}
      \begin{pmatrix}
        t\\
        s
      \end{pmatrix} - 2\begin{pmatrix}
        d & e
      \end{pmatrix}\begin{pmatrix}
        t\\
        s
      \end{pmatrix} + \bar{f},
 \end{align}
 with the following variables
 \begin{align}
   \Sigma^{-1} \triangleq
   \begin{pmatrix}
     a & b\\
     b & c
   \end{pmatrix} &=         
       \begin{pmatrix}
         \mathbf{w}_i^{\Transp}\mathbf{Vw}_i & -\mathbf{w}_i^{\Transp}\mathbf{Vw}_j\\
         -\mathbf{w}_j^{\Transp}\mathbf{Vw}_i & \mathbf{w}_j^{\Transp}\mathbf{Vw}_j
       \end{pmatrix},\\
   d &= (\mathbf{p}_i - \mathbf{w}_j)^{\Transp} \mathbf{V} \mathbf{w}_i,\\
   e &= -(\mathbf{p}_i - \mathbf{p}_j)^{\Transp} \mathbf{V} \mathbf{w}_j,\\
   \bar{f} &= (\mathbf{p}_i - \mathbf{p}_j)^{\Transp}\mathbf{V}(\mathbf{p}_i - \mathbf{p}_j).
 \end{align}
\end{subequations}

When the matrix $\Sigma^{-1}$ is invertible, then it is possible to
reformulate the integral and employ integration methods for bivariate
normal distributions (see e.g. \cite{genz2004numerical} and the references
therein). 

However, when $0<|ac-b^2| < \epsilon$ for some small value $\epsilon >
0$, then this approach fails to reliably deliver accurate
solutions. This problem is most notable when $ac-b^2=0$, in which case
$\Sigma$ is not invertible and the bivariate approach cannot be used. There is an alternate approach in this particular case, but the user is now faced with a choice to numerically determine the appropriate threshold to decide between the solution methods. This is most notably an issue for lines that are close to co-linear. Importantly, the
case where $ac-b^2=0$ occurs frequently, e.g. along the diagonal
entries of the covariance matrix.

Transformation of the problem into the Bivariate Normal is presented in the Appendix of \cite{hennig2012quasi}. However the non-invertible case is not explicitly dealt with, therefore we have provided the details for applying this approach and the necessary cases in Appendix~\ref{app:bivariate_normal_method}.

\section{Efficient Computation of the Double Line Integral} 
\label{sec:method}
Here, a method to accurately evaluate the double line integral \eqref{eq:doubleInt} in a computationally efficient manner is presented. 
The integral can be rewritten as
\begin{subequations}
\begin{equation}
\begin{split}
  \mathbf{K}_{ij} &= \norm{\mathbf{w}_i}\norm{\mathbf{w}_j}\int\limits_0^1\int\limits_0^1\exp\Bigg(-\frac{1}{2}\big(a -bs +ct- dst + et^2 + fs^2\big)\Bigg)\,\mathrm{d}t\mathrm{d}s,
\end{split}
\end{equation}
where
\begin{equation}
\begin{matrix}
        a = \mathbf{u}_{ij}^T\mathbf{V}\mathbf{u}_{ij}, & b = 2\mathbf{u}_{ij}^T\mathbf{V}\mathbf{w}_j, & c = 2\mathbf{u}_{ij}^T\mathbf{V}\mathbf{w}_i, \\
        d = 2\mathbf{w}_j^T\mathbf{V}\mathbf{w}_i, & e = \mathbf{w}_i^T\mathbf{V}\mathbf{w}_i, & f = \mathbf{w}_j^T\mathbf{V}\mathbf{w}_j. \\
\end{matrix}
\end{equation}
\end{subequations}

The error function can be used to provide a solution to the integral over $t$;
\small
\begin{equation}\label{eq:sol1}
\begin{split}
    \mathbf{K}_{ij} &=\norm{\mathbf{w}_i}\norm{\mathbf{w}_j}\sqrt{\frac{\pi}{2e}}\int_0^1\Gamma_1(s)\Gamma_2(s)\,\mathrm{d}s,
\end{split}
\end{equation}
where
\begin{equation}
  \begin{split}
    \Gamma_1(s) &= \exp\left(\frac{1}{2}\left(bs - a - fs^2 + \frac{(c-ds)^2}{4e}\right)\right) \\
    \Gamma_2(s) &=\text{erf}\left(\frac{c-ds+2e}{2\sqrt{2e}}\right)-\text{erf}\left(\frac{c-ds}{2\sqrt{2e}}\right).
  \end{split}
\end{equation}
\normalsize

For numerical reasons the exponent should not be split into a product of $\exp(-a)$ and $\exp\left(\frac{1}{2}\left(bs - fs^2 + \frac{(c-ds)^2}{4e}\right)\right)$. The reason is that when $\norm{\mathbf{V}\mathbf{u}_{ij}}$ is large, $\exp(-a) \to 0$ while $\exp\left(\frac{1}{2}\left(bs - fs^2 + \frac{(c-ds)^2}{4e}\right)\right)\to \infty$ and rounding errors become a problem.

Numerical methods can then be used to evaluate the remaining integral. 
Care should be taken when considering the following two cases:

\textbf{Case 1:} Either $\norm{\mathbf{w}_i} = 0$ or $\norm{\mathbf{w}_j} = 0$.  Then the problem reduces to a single line integral and two sub cases should be considered in implementation.
\begin{itemize}
\item[]\textbf{Case 1a:} $\norm{\mathbf{w}_i} = 0$ and $\norm{\mathbf{w}_j} > 0$. Then Equation~\eqref{eq:sol1} is numerically unstable and two approaches can be taken. Either the solution to the single integral \eqref{eq:solsingle2} can be used, which requires the user to determine a numerical threshold to decide between the two cases. 
Alternatively, the order of the integrals can be swapped, and Equation~\eqref{eq:sol1} can be applied. 
\item[]\textbf{Case 1b:} $\norm{\mathbf{w}_i} > 0$ and $\norm{\mathbf{w}_j} = 0$. Equation~\eqref{eq:sol1} is numerically stable, and hence the user can either ignore this case or define a numerical tolerance below which to use the solution to the single integral \eqref{eq:solsingle2}.
\end{itemize}
\textbf{Case 2:} $\norm{\mathbf{w}_i} = 0$ and $\norm{\mathbf{w}_j} = 0$. Then Equation \eqref{eq:sol1} is numerically unstable and the solution is instead given by
\begin{equation}
  \mathbf{K}_{ij} = \norm{\mathbf{w}_i}\norm{\mathbf{w}_j}\exp\left(-\frac{1}{2}a\right) = 0.
\end{equation}
Psudocode is provided in Algorithm~\ref{alg:psudo_code} and a Matlab mex function implemented in \texttt{C} is provided at \cite{githubcode}. In the implementation and psuedocode, the order of the lines is swapped whenever $\norm{\mathbf{Vw}_j} > \norm{\mathbf{Vw}_i}$ as this means the $\erf()$ function is used to evaluate the larger interval and the numerical integration the shorter interval; which was found to require less function evaluations. The C implementation utilises the BLAS functions \texttt{dgemm}, \texttt{ddot}, and \texttt{dnrm2} \cite{dongarra1990set}, and the GNU Scientific Library's non-adaptive Gauss Konrod function \texttt{gsl\textunderscore integration\textunderscore qng} \cite{gough2009gnu} with relative and absolute error tolerance set to square root of double precision epsilon.

\begin{algorithm}[htb]
      \caption{Psuedocode for evaluation of Equation~\ref{eq:doubleInt}}
      \label{alg:psudo_code}
        \begin{algorithmic}[1]
          \Procedure{Double Integral}{$\mathbf{u}_{ij}$, $\mathbf{w}_i$, $\mathbf{w}_j$, $\mathbf{V}$, $\epsilon$}
          \If{$||\mathbf{V}\mathbf{w}_i || < \epsilon $ \&\& $ ||\mathbf{V}\mathbf{w}_2 || <  \epsilon$} 
            \State $\mathbf{K}_{ij} := \norm{\mathbf{w}_i}\norm{\mathbf{w}_j}\exp\left(-\frac{1}{2}\mathbf{u}_{ij}^T\mathbf{V}\mathbf{u}_{ij}\right)$
          \Else
            \If{$\norm{\mathbf{Vw}_i} < \norm{\mathbf{Vw}_j}$}
            \State $a := \mathbf{u}_{ij}^T\mathbf{V}\mathbf{u}_{ij}$,\quad \quad \  $b := -2\mathbf{u}_{ij}^T\mathbf{V}\mathbf{w}_i$, 
            \State  $c := -2\mathbf{u}_{ij}^T\mathbf{V}\mathbf{w}_j$,\quad  $d := 2\mathbf{w}_i^T\mathbf{V}\mathbf{w}_j$
            \State $e := \mathbf{w}_j^T\mathbf{V}\mathbf{w}_j$, \quad \ \ \ $f := \mathbf{w}_i^T\mathbf{V}\mathbf{w}_i$ 
            \Else
              \State $a := \mathbf{u}_{ij}^T\mathbf{V}\mathbf{u}_{ij}$,\quad \ $b := 2\mathbf{u}_{ij}^T\mathbf{V}\mathbf{w}_j$
              \State $c := 2\mathbf{u}_{ij}^T\mathbf{V}\mathbf{w}_i$,\quad $d := 2\mathbf{w}_j^T\mathbf{V}\mathbf{w}_i$, 
              \State $e := \mathbf{w}_i^T\mathbf{V}\mathbf{w}_i$, \quad $f := \mathbf{w}_j^T\mathbf{V}\mathbf{w}_j$
            \EndIf
            \State Evaluate $\mathbf{K}_{ij}$ by applying a 1D numerical integration technique to equation \eqref{eq:sol1}
          \EndIf
          \State \textbf{return} $\mathbf{K}_{ij}$ 
          \EndProcedure
        \end{algorithmic}
    \end{algorithm}


\section{Results and Analysis} 
\label{sec:analysis}
The performance of the methods was evaluated for several sets of double line integrals; containing 10,000 pairs each and with input dimension $m=6$. The sets were chosen to test a variety of cases and in particular to evaluate the performance on the transition between the corner cases. The following sets were used:
\begin{itemize}
  \item[] \textbf{Set 1:} A standard set; $\mathbf{V} = \mathbf{I}$, $\mathbf{w}_{i,k} \sim \mathcal{U}(0,1)$, $\mathbf{w}_{j,k} \sim \mathcal{U}(0,1)$, and $\mathbf{u}_{ij,k} \sim \mathcal{U}(0,1)$ for $k=1,\dots,m$,
  \item[] \textbf{Set 2:} Almost colinear lines; $\mathbf{V} =\mathbf{I}$, $\mathbf{w}_{i,k} \sim \mathcal{U}(0,1)$, $\mathbf{w}_{j,k} \sim \mathbf{w}+\mathcal{U}(0,1\times10^{-8})$, and $\mathbf{u}_{ij,k} \sim \mathcal{U}(0,1)$ for $k=1,\dots,m$,
  \item[] \textbf{Set 3:} Randomly selected diagonal scaling matrix; $V_{kk} \sim  \mathcal{U}(0,1)$, $\mathbf{w}_{i,k} \sim  \mathcal{U}(0,1)$, $\mathbf{w}_{j,k} \sim  \mathcal{U}(0,1)$, and $\mathbf{u}_{ij,k} \sim \mathcal{U}(0,1)$ for $k=1,\dots,m$,
  \item[] \textbf{Set 4:} Reduced to single integral; $\mathbf{V} = \mathbf{I}$, $\mathbf{w}_i = \mathbf{0}$, $\mathbf{w}_{j,k} \sim \mathcal{U}(0,1)$, and $\mathbf{u}_{ij,k} \sim \mathcal{U}(0,1)$ for $k=1,\dots,m$,
  \item[] \textbf{Set 5:} Nicely scaled (large integral intervals and small random scaling); $V_{kk} \sim \mathcal{U}(0,0.01)$, $\mathbf{w}_{i,k} \sim \mathcal{U}(0,10)$, $\mathbf{w}_{j,k} \sim \mathcal{U}(0,10)$, and $\mathbf{u}_{ij,k} \sim \mathcal{U}(0,10)$ for $k=1,\dots,m$,
  \item[] \textbf{Set 6}: Poorly scaled (large integral intervals and large random scaling); $V_{kk} \sim \mathcal{U}(0,10)$, $\mathbf{w}_{i,k} \sim \mathcal{U}(0,10)$, $\mathbf{w}_{j,k} \sim \mathcal{U}(0,10)$, and $\mathbf{u}_{ij,k} \sim \mathcal{U}(0,10)$ for $k=1,\dots,m$,
  \item[] \textbf{Set 7:}: Almost reduced to single integral; $\mathbf{V} = \mathbf{I}$, $\mathbf{w}_{i,k} \sim \mathcal{U}(0,1\times10^{-8})$, $\mathbf{w}_{j,k} \sim \mathcal{U}(0,1)$, and $\mathbf{u}_{ij,k} \sim \mathcal{U}(0,1)$ for $k=1,\dots,m$,
  \item[] \textbf{Set 8:} Almost reduced to no integral; $\mathbf{V} = \mathbf{I}$, $\mathbf{w}_{i,k} \sim \mathcal{U}(0,1\times10^{-8})$, $\mathbf{w}_{j,k} \sim \mathcal{U}(0,1\times10^{-8})$, and $\mathbf{u}_{ij,k} \sim \mathcal{U}(0,1)$ for $k=1,\dots,m$.
\end{itemize}
Here, the notation $\mathbf{v}_{i,k}~\sim\mathcal{U}(a,b)$ denotes that the $k^\text{th}$ element of the vector $\mathbf{v}_i$ is distributed uniformly between $a$ and $b$.

Errors were calculated by comparison to Matlab's \texttt{Integral2} using absolute and relative error tolerance of double precision epsilon. The mean magnitude of the error for each set are recorded in Table~\ref{tab:mean_errors}. The computation times required by each method were consistent across the cases and so the average computation times are recorded in Table~\ref{tab:comp_times}.

\newcommand{\ra}[1]{\renewcommand{\arraystretch}{#1}}
\begin{table*}[t]
\ra{1.3}
\centering
\scalebox{0.8}{
\begin{tabular}{@{}lrrrrrrrr@{}}\toprule
Method & Set 1 & Set 2 & Set 3& Set 4 & Set 5& Set 6 & Set 7 & Set 8  \\ 
\midrule
Simpson's Rule\\
\hspace{5mm}$p=10$ &$3.66\times10^{-6}$&$3.49\times10^{-6}$&$1.62\times10^{-6}$&$0$&$1.63\times10^{-4}$&$8.93\times10^{-5}$&$3.14\times10^{-14}$&$2.27\times10^{-32}$\\ 
\hspace{5mm}$p=100$&$3.62\times10^{-10}$&$3.44\times10^{-10}$&$1.61\times10^{-10}$&$0$&$1.61\times10^{-8}$&$7.08\times10^{-10}$&$3.12\times10^{-18}$&$9.07\times10^{-32}$\\ 
\hspace{5mm}$p=200$ &$2.23\times10^{-11}$&$2.15\times10^{-11}$&$1.00\times12^{-11}$&$0$&$1.01\times10^{-9}$&$4.39\times10^{-11}$&$1.95\times10^{-19}$&$1.27\times10^{-31}$\\ \addlinespace[0.5mm]
\begin{tabular}[c]{@{}l@{}}Bivariate Normal\end{tabular}     &$2.06\times10^{-15}$& $1.84\times10^{-1}$&$5.03\times10^{-15}$&$0$&$4.81\times10^{-13}$&$9.30\times10^{-17}$&$3.26\times10^{-17}$&$1.58\times10^{-25}$\\ 
Proposed &$1.80\times10^{-15}$&$6.39\times10^{-16}$&$4.35\times10^{-15}$&$0$&$4.15\times10^{-13}$&$8.32\times10^{-14}$&$9.56\times10^{-24}$&$2.10\times10^{-25}$\\
\bottomrule
\end{tabular}}
\caption{Mean magnitude of the error for each set. Errors are computed with respect to Matlab's \texttt{Integral2} function.}
\label{tab:mean_errors}
\end{table*}

\begin{table}[]
\centering
\begin{tabular}{@{}lr@{}} \toprule
Method & Time ($\times10^{-5}$s)\\ \midrule
Matlab Integral2& 149.38* \\ 
Simpson's Rule \\
\hspace{5mm} $p=10$ & $1.24$\\ 
\hspace{5mm} $p=100$& $14.54$\\ 
\hspace{5mm} $p=200$ & $54.86$\\ \addlinespace[0.5mm]
Bivariate Normal & $6.27$\\ 
Proposed & $1.21$\\ \bottomrule
\end{tabular}
\caption{Average computation time across all sets for each method. Note* the reported time is for \texttt{Integral2} with absolute and relative tolerance of double precision epsilon, using the default tolerances reduces the computation time to $0.0011687\si{s}$.}
\label{tab:comp_times}
\end{table}

A 2D Simpson's rule was implemented with $p$ sub intervals along $t$ and $s$; results are shown for $p=10$, $p=100$ and $p=200$. This 2D Simpson's rule gives the lowest error for Set~8 where the integrand is close to a constant function and is quite fast a choice of small $p$. Sets 1 through 6 it show relatively large errors with the errors decreasing as $p$ is increased. However, even with $p=200$ the errors are still larger than for the Bivariate Normal method and the proposed method and the computation times are an order of magnitude greater. 

\begin{figure}[!ht]
    \centering
    \includegraphics[width=0.5\linewidth]{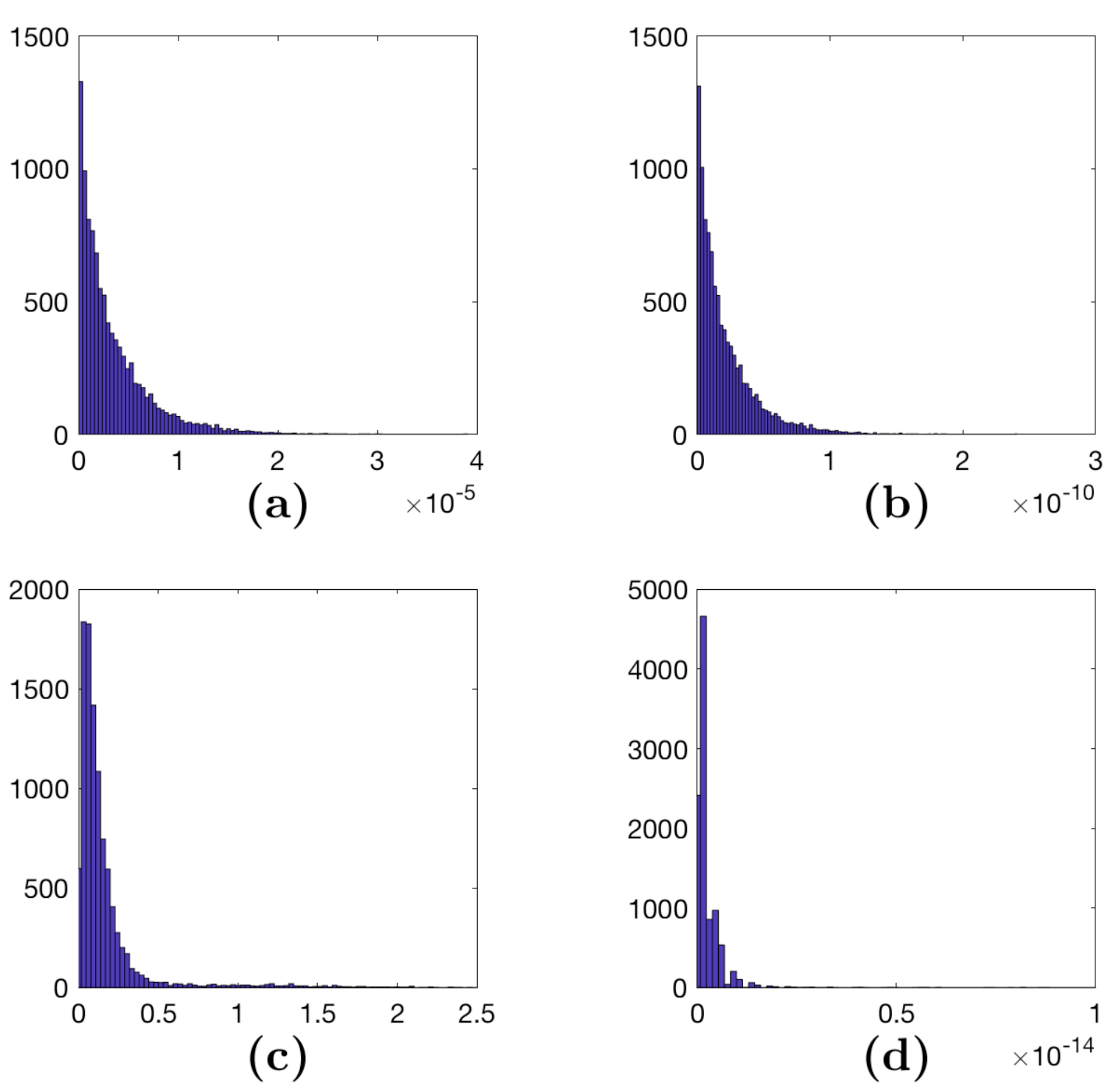}
    \caption{Histogram of the error magnitudes for evaluating the double line integrals in set 2 (the almost co-linear case); (a) Simpson's rule with $p=10$, (b) Simpson's rule with $p=200$, (c) Bivariate Normal method, (d) our proposed method.}
    \label{fig:error_hist}
\end{figure}

The Bivariate Normal method has acceptable small errors across all sets excluding set 2. The Bivariate normal method required a different algorithm for the co-linear and non co-linear cases and Set 2 lies on the transition between these cases. Figure~\ref{fig:error_hist} provides a histogram of the error magnitudes for Set~2 as computed by each method. The histogram shows that the Bivariate normal method can result in large errors which would be prohibitive for many applications. 

Our proposed method has acceptable errors across all sets. The worst errors, of $1\times10^{-13}$ and $1\times10^{-14}$ for Set~5 and Set~6 respectively, could potentially be improved by using an adaptive integration method that may be more suited to the scaling in these sets. The average computation time is the smallest reported.

\section{Conclusion} 
\label{sec:conslucion}
In this paper we have focussed on the specific problem of evaluating a double line integral over the squared-exponential covariance function. This problem arises in several applications of Gaussian process regression. Existing approaches to this problem and their advantages and disadvantages were outlined. 
An alternative method was proposed and its performance compared to the existing methods was evaluated for a number of sets of line integrals corresponding to both common cases and corner cases. 
The comparison shows the proposed method is computationally efficient while providing acceptable accuracy across all tested cases.

\section*{Acknowledgements}                               
This research was financially supported by the Swedish Foundation for Strategic Research (SSF) via the project \emph{ASSEMBLE} (contract number: RIT15-0012) and by the Swedish Research Council via the projects \emph{Learning flexible models for nonlinear dynamics} (contract number: 2017-03807) and \emph{NewLEADS - New Directions in Learning Dynamical Systems} (contract number: 621-2016-06079).

\appendix
\section{Bivariate normal method} 
\label{app:bivariate_normal_method}
In this appendix we show how to compute the double integral in \eqref{eq:doubleInt} by reformulating the problem as the double integral of a Bivariate normal distribution.
The integral can be written as
\begin{subequations}
  \begin{align}
    I_2 = \int_{0}^{1} \int_{0}^{1} e^{-\frac{1}{2}f(t,s)}
    \myd t \myd s,
  \end{align}
  where the function $f(s, t)$ is given by
 \begin{align}
   \notag
   f(t, s) &= (\mathbf{p}_i - \mathbf{p}_j + \mathbf{w}_{i}t- \mathbf{w}_{j}s)^\Transp \mathbf{V}(\mathbf{p}_i - \mathbf{p}_j + \mathbf{w}_{i}t - \mathbf{w}_{j}s)\\
      &= \begin{pmatrix}
        t\\
        s
      \end{pmatrix}^{\Transp}
   \begin{pmatrix}
     a & b\\
     b & c
   \end{pmatrix}
      \begin{pmatrix}
        t\\
        s
      \end{pmatrix} - 2\begin{pmatrix}
        d & e
      \end{pmatrix}\begin{pmatrix}
        t\\
        s
      \end{pmatrix} + \bar{f},
 \end{align}
  with the following variables
  \begin{align}
    \begin{pmatrix}
      a & b\\
      b & c
    \end{pmatrix} &=    
                    \begin{pmatrix}
                      \mathbf{w}_i^{\Transp}V\mathbf{w}_i & -\mathbf{w}_i^{\Transp}V\mathbf{w}_j\\
                      -\mathbf{w}_j^{\Transp}V\mathbf{w}_i & \mathbf{w}_j^{\Transp}V\mathbf{w}_j
                    \end{pmatrix},\\
    d &= -(\mathbf{p}_i - \mathbf{p}_j)^{\Transp} V \mathbf{w}_i,\\
    e &= (\mathbf{p}_i - \mathbf{p}_j)^{\Transp} V \mathbf{w}_j,\\
    \bar{f} &= (\mathbf{p}_i - \mathbf{p}_j)^{\Transp}V(\mathbf{p}_i - \mathbf{p}_j).
  \end{align}
\end{subequations}
There are two cases that need to be considered separately as the
solutions to each case do not commute. The reason for needing two
cases is that when $\mathbf{w}_i$ is colinear with $\mathbf{w}_j$ (as occurs along the
block diagonal elements of the $K_{\ell_k, \ell_k}$ matrix), then the above
matrix involving $a,b,c$ is not invertible. 

\subsection{Double integral with $ac-b^2=0$}
This case can occur in four distinct ways.

\textbf{Case 1:} $\|\mathbf{w}_i\| = 0$ and $\|\mathbf{w}_j\| > 0$. Then $a=b=d=0$
and $f(t,s) = \bar{f} - 2es + cs^2$ and we can use \eqref{eq:solsingle1} and \eqref{eq:solsingle2} to solve this integral. 

\textbf{Case 2:} $\|\mathbf{w}_i\| > 0$ and $\|\mathbf{w}_j\| = 0$. Then $c=b=e=0$
and $f(t,s) = \bar{f} + 2 d t + at^2$ and we can again use \eqref{eq:solsingle1} and \eqref{eq:solsingle2} to solve this integral.

\textbf{Case 3:} $\|\mathbf{w}_j\| = 0$ and $\|\mathbf{w}_i\| = 0$. Then $a=b=c=d=e=0$
and $f(t,s) = \bar{f}$ and the solution is $I_2 = e^{-\frac{\bar{f}}{2}}.$

\textbf{Case 4:} $\|\mathbf{w}_i\| > 0$ and $\|\mathbf{w}_j\| > 0$. Since we are already
in the case where $ac-b^2 = 0$, and $a > 0$ and $c > 0$ then it
follows that $\mathbf{w}_i = \beta \mathbf{w}_j$ for some $\beta \neq 0$ since 
\begin{align*}
  ac - b^2 &= (\mathbf{w}_i^\Transp V \mathbf{w}_i) (\mathbf{w}_j^\Transp V \mathbf{w}_j) - (\mathbf{w}_i^\Transp V
  \mathbf{w}_j)^2,\\
  &=(\mathbf{w}_i^\Transp V \mathbf{w}_i) (\beta^2 \mathbf{w}_i^\Transp V \mathbf{w}_i) - (\beta \mathbf{w}_i^\Transp V
  \mathbf{w}_i)^2,\\
  &= \beta^2 (\mathbf{w}_i^\Transp V \mathbf{w}_i)^2 - \beta^2 (\mathbf{w}_i^\Transp V \mathbf{w}_i)^2\\
  &= 0 
\end{align*}
as claimed. Therefore, by defining $\beta$ as
\begin{align}
  \beta \triangleq \frac{\mathbf{w}_i^\Transp V \mathbf{w}_j}{\mathbf{w}_i^\Transp V \mathbf{w}_i}
\end{align}
then $f(t,s)$ can be expressed as 
\begin{align*}
  f(t,s) &= f + 2t(\mathbf{p}_i-\mathbf{p}_j)^\Transp V \mathbf{w}_i - 2 \beta s
  (\mathbf{p}_i-\mathbf{p}_j)^\Transp V \mathbf{w}_i+ t^2 (\mathbf{w}_i^TV\mathbf{w}_i) -2\beta t s (\mathbf{w}_i^TV\mathbf{w}_i) 
  + \beta^2 s^2 (\mathbf{w}_i^TV\mathbf{w}_i).
\end{align*}
If we make a change of variables for $s$ to
\begin{align*}
  \bar{s} = \beta s \quad \implies \quad
  f(t,\bar{s}) = f + a(t^2 -2 t \bar{s}
  + s^2 )
\end{align*}
and the integral becomes
\begin{align*}
  I_2 &= \frac{e^{-f/2}}{\beta}\int_0^\beta \int_0^1 e^{-\frac{a}{2}(t^2 -2 t \bar{s}
  + \bar{s}^2)} \myd t \myd \bar{s}\\
&= \frac{e^{-f/2}}{\beta}\int_0^\beta \int_0^1 e^{-\frac{a}{2}(t - \bar{s})^2} \myd t \myd \bar{s}
\end{align*}
By a change of variables $\phi = \sqrt{\frac{a}{2}}(t - \bar{s})$ then 
\begin{align}
  I_2 &= \frac{\sqrt{2}e^{-f/2}}{\beta\sqrt{a}}\int_0^\beta
        \int_{-\bar{s}\sqrt{a/2}}^{(1-\bar{s})\sqrt{a/2}}
        e^{-\phi^2} \myd \phi \myd \bar{s}\\
  &= \frac{\sqrt{\pi}\sqrt{2}e^{-f/2}}{2\beta\sqrt{a}} \int_0^\beta
    \left ( \erf((1-\bar{s})\sqrt{a/2}) -
    \erf(-\bar{s}\sqrt{a/2}) \right ) \myd \bar{s}
\end{align}
where the second equality stems from the definition of the error
function 
\begin{align}
 \erf(x) = \frac{2}{\sqrt{\pi}}\int_{0}^x e^{-u^2} \myd u. 
\end{align}
Next we can exploit the fact that the integral of the error function
is given by (using integration by parts)
\begin{align}
  \int_a^b \erf(x) dx = b\, \erf(b) + \frac{e^{-b^2}}{\sqrt{\pi}} - a\, \erf(a) - \frac{e^{-a^2}}{\sqrt{\pi}}.
\end{align}
To apply this to $I_2$ it is helpful to split the integral component
via
\begin{align}
  I_2 &= \frac{\sqrt{\pi}\sqrt{2}e^{-f/2}}{2\beta\sqrt{a}} \left (
        I_3 - I_4 \right ),\\
  I_3 &\triangleq \int_0^\beta \erf((1-\bar{s})\sqrt{a/2})\myd \bar{s}\\
  I_4 &\triangleq \int_0^\beta \erf(-\bar{s}\sqrt{a/2}) \myd \bar{s}
\end{align}
By change of variables $\zeta = (1-\bar{s})\sqrt{a/2}$ then $I_3$
becomes
\begin{align*}
  I_3 &= \sqrt{\frac{2}{a}} \int_{(1-\beta)\sqrt{a/2}}^{\sqrt{a/2}}
  \erf(\zeta) \myd \zeta\\
      &= \sqrt{\frac{2}{a}} \Bigg [ \sqrt{\frac{a}{2}} \erf\left(\sqrt{\frac{a}{2}}\right) +
        \frac{e^{-\frac{a^2}{2}}}{\sqrt{\pi}}- 
        (1-\beta)\sqrt{\frac{a}{2}} \erf\left((1-\beta)\sqrt{\frac{a}{2}}\right) -
        \frac{e^{-(1-\beta)^2\frac{a^2}{2}}}{\sqrt{\pi}} \Bigg ]
\end{align*}
And similarly for $I_4$ with change of variables $\psi = -\bar{s}
\sqrt{a/2}$ to arrive at
\begin{align}
  I_4 &= \sqrt{\frac{2}{a}} \int_{-\beta\sqrt{a/2}}^{0}
  \erf(\psi) \myd \psi\\
  &= \sqrt{\frac{2}{a}} \Bigg [ \frac{1}{\sqrt{\pi}} + \beta\sqrt{\frac{a}{2}} \erf\left(-\beta\sqrt{\frac{a}{2}}\right) -
        \frac{e^{-\beta^2\frac{a^2}{2}}}{\sqrt{\pi}} \Bigg ]
\end{align}
\subsection{Double integral with $ac-b^2>0$}
By defining 
\begin{align*}
  z &= \begin{pmatrix}
    t\\
    s
  \end{pmatrix}, \quad 
  \Sigma^{-1} = \begin{pmatrix}
      a & b\\
      b & c
    \end{pmatrix}, \quad 
    \nu = \begin{pmatrix}
    d\\
    e
  \end{pmatrix},\quad
  \mu = \Sigma \nu\\
  g(z) &\triangleq (z-\mu)^\Transp \Sigma^{-1} (z-\mu), \qquad
  h \triangleq f - \mu^{\Transp} \Sigma^{-1} \mu
\end{align*}
we have that
\begin{align}
    f(t, s) &= g(z) + h
\end{align}
Therefore,
\begin{align}
  e^{-\frac{1}{2}f(t, s)} &= e^{-\frac{1}{2}g(z) -\frac{h}{2}}
  = 2\pi e^{-\frac{h}{2}} \sqrt{\det \Sigma} \Npdf{z}{\mu}{\Sigma}
\end{align}
where $\Npdf{z}{\mu}{\Sigma}$ denotes the probability density function
for the normal distributed random variable~$z$, with mean value~$\mu$
and covariance~$\Sigma$. With the change of variables
$\bar{z} = z-\mu$ the integral can be expressed as
\begin{align}
  \label{eq:I2b}
  I_2 = 2\pi e^{-\frac{h}{2}} \sqrt{\det \Sigma}\int_{-\mu_1}^{1 - \mu_1} \int_{-\mu_2}^{1 - \mu_2} 
  \Npdf{\bar{z}}{0}{\Sigma}
               \myd \bar{z}_1 \myd \bar{z}_2,
\end{align}
Note that the density in~\eqref{eq:I2b} is a bivariate normal
distribution, which means that it can be written as
  \begin{align*}
\Npdf{\bar{z}}{0}{\Sigma} 
    &= \frac{1}{2\pi\sqrt{1-\rho^2}\sqrt{c_{11}c_{22}}}
      \exp\left(-\frac{\alpha(\bar{z})}{2(1-\rho^2)}\right)
  \end{align*}
  where 
  \begin{align*}
    &\pmat{c_{11} & c_{12} \\ c_{21} & c_{22}} \triangleq \Sigma =
                                      \pmat{a & -b\\-b & c}^{-1} = 
                                      \frac{1}{ac-b^2}\pmat{c & b\\b
    & a}\\
    \alpha(\bar{z}) &=  \left(\frac{\bar{z}_1}{\sqrt{c_{11}}}\right)^2 - 2\rho\left(\frac{\bar{z}_1}{\sqrt{c_{11}}}\right)\left(\frac{\bar{z}_2}{\sqrt{c_{22}}}\right)
             + \left(\frac{\bar{z}_2}{\sqrt{c_{22}}}\right)^2\\
    \rho &= \frac{c_{12}}{\sqrt{c_{11}c_{22}}}.
  \end{align*}
By change of variables $\widetilde{z}_1 =
\frac{\bar{z}_1}{\sqrt{c_{11}}}$ and $\widetilde{z}_2 = \frac{\bar{z}_2}{\sqrt{c_{22}}}$
the resulting integral as
\begin{subequations}
  \label{eq:App:LI:d2}
  \begin{align}
    \label{eq:App:LI:d2a}
    I_2 = 2\pi e^{-\frac{h}{2}} \sqrt{\det \Sigma}\int_{-\frac{\mu_1}{\sqrt{c_{11}}}}^{\frac{1 - \mu_1}{\sqrt{c_{11}}}}
    \int_{-\frac{\mu_2}{\sqrt{c_{22}}}}^{\frac{1 - \mu_2}{\sqrt{c_{22}}}}
    \beta(\rho)
    \myd \widetilde{z}_1 \myd \widetilde{z}_2.
  \end{align}
  where 
  \begin{align}
    \beta(\rho) = \frac{1}{2\pi\sqrt{1-\rho^2}}\exp\left(-\frac{\widetilde{z}_1^2 - 2\rho\widetilde{z}_1\widetilde{z}_2 + \widetilde{s}_2^2}
    {2(1-\rho^2)}\right).
  \end{align}
\end{subequations}
The integral~\eqref{eq:App:LI:d2a} has been widely studied in the
literature, see e.g. \cite{genz2004numerical} and the references
therein. There is efficient code
available\footnote{\url{www.math.wsu.edu/faculty/genz/software/matlab/bvn.m}}
to compute this integral.

\bibliographystyle{plain}        
\bibliography{References}           

\end{document}